\documentclass[runningheads]{llncs}
\usepackage[T1]{fontenc}
\usepackage{graphicx}
\usepackage{booktabs} 
\usepackage{comment}
\usepackage{xcolor}

\usepackage{tablefootnote}

\usepackage{enumitem,amssymb}
\usepackage{amsmath}
\newlist{todolist}{itemize}{2}
\setlist[todolist]{label=$\square$}
\usepackage{pifont}

\begin{document}
\title{EchoNarrator: Generating natural text explanations for ejection fraction predictions}
\titlerunning{EchoNarrator: Generating natural text explanations}
\author{Sarina Thomas \inst{1,2} \and 
Qing Cao\inst{3} \and 
Anna Novikova\inst{4} \and \newline 
Daria Kulikova\inst{4} \and 
Guy Ben-Yosef \inst{5,\dagger} 
}

\authorrunning{S. Thomas et al.}
%
\institute{University of Oslo, Oslo, Norway \and
Department of Cardiovascular Ultrasound, GE Healthcare, Oslo, Norway \and
GE Healthcare, Wuxi, China \and
GE Healthcare, Kharxiv, Ukraine \and
GE HealthCare Technology \& Innovation Center, Niskayuna, New York, USA \\ $^{\dagger}$ corresponding author \\ \email{guy.ben-yosef@gehealthcare.com} }

\maketitle 
\begin{abstract}
Ejection fraction (EF) of the left ventricle (LV) is considered as one of the most important measurements for diagnosing acute heart failure and can be estimated during cardiac ultrasound acquisition. While recent successes in deep learning research successfully estimate EF values, the proposed models often lack an explanation for the prediction. However, providing clear and intuitive explanations for clinical measurement predictions would increase the trust of cardiologists in these models.
In this paper, we explore predicting EF measurements with Natural Language Explanation (NLE). We propose a model that in a single forward pass combines estimation of the LV contour over multiple frames, together with a set of modules and routines for computing various motion and shape attributes that are associated with ejection fraction. It then feeds the attributes into a large language model to generate text that helps to explain the network's outcome in a human-like manner. We provide experimental evaluation of our explanatory output, as well as EF prediction, and show that our model can provide EF comparable to state-of-the-art together with meaningful and accurate natural language explanation to the prediction. The project page can be found at https://github.com/guybenyosef/EchoNarrator . 

\end{abstract}

\section{Introduction}
\label{sec:intro}

The release of the extensive Dynamic EchoNet echocardiography dataset~\cite{ouyang2020video} has accelerated the adoption of deep learning models for ejection fraction (EF) prediction and left ventricle (LV) contour delineation. Several innovative approaches have been introduced, including LV segmentation \cite{ouyang2020video,Weihang2023,Meng2024}, direct video regression \cite{ouyang2020video,kazemi2020deep,reynaud2021ultrasound,rand2022miccai}, graph- and keypoints-based models \cite{Mokhtari2022miccai,ours2022miccai}, and attention-based models \cite{Mokhtari2023} - Their potential is somewhat diminished by a common shortfall: the lack of clinically meaningful explanations for the predicted EF. 
Explainability of visual deep learning models is often linked with activation maps such as Class Activation Mapping (CAM)~\cite{zhou2016learning} and Grad-CAM~\cite{selvaraju2017grad}, which associate image regions with their contribution to the prediction. Although useful in certain contexts, these methods often fall short in medical imaging, where they may only highlight obvious regions such as the LV to predict EF, providing clinically correct but not meaningful information (an interesting example of this phenomenon was shown in~\cite{Mokhtari2023}). To overcome these limitations and improve the explainability with human-like text, a novel approach has been developed within the subfield of {\em Natural Language Explanations} (NLE). This approach leverages advancements in vision-language and language models to generate text explanations that accompany model outcomes, providing context and clarity that activation maps cannot provide. NLE models include the generation of text explanations based on object attributes~\cite{hendricks2016generating}, models for associating text explanations with image regions~\cite{hendricks2018grounding}, and language models such as GPT2 \cite{sammani2022nlx,kayser2022explaining} and GPT3 \cite{sammani2023iccvw}. Considering that abnormal EF values are often linked to visible changes in the LV, providing explanations based on those visual cues shall enhance cardiologists' confidence in the deep learning predictions. 
Inspired by attribute-based NLE strategies (e.g., ~\cite{hendricks2016generating,hendricks2018grounding}), where explanations are generated based on a predefined set of attributes, we create attributes that influence EF predictions, such as wall thickening in the interventricular septum (hereafter referred to as bulge), regional wall motion abnormalities, and foreshortening due to acquisition. Furthermore, we incorporate Large Language Models (LLMs) into our pipeline, utilizing models like LLaMA~\cite{touvron2023llama} as the final step to synthesize smooth and coherent explanations. 
Our innovation harnesses the capabilities of LLMs to assimilate the estimation of relevant LV attributes, generating text explanations that are informative and aligned with clinical practices. This method marks a significant step toward developing an AI assistant capable of engaging with clinicians through human-like language. 

Our paper presents three major contributions to the field of cardiovascular ultrasound analysis and interpretation:
\begin{enumerate}
    \item[(1)]{{\em Novel NLE Model for EF Prediction:} We introduce the first NLE model tailored for EF prediction in cardiovascular ultrasound. This model synergizes the analytical depth of modern LLMs with spatiotemporal analysis of geometric features, setting a new benchmark for accuracy and explainability.}
    \item[(2)]{{\em Self-Instruction Training Method:} We develop a novel training approach for the LLaMA model, utilizing GPT-4 to augment explanation examples. By releasing a dataset of approx. 800 self-instructions, we lay the groundwork for future advancements in training LLMs for echocardiography-related tasks.}
    \item[(3)]{{\em Evaluation Metrics for Explanation Output:} Our research extends into the development and application of evaluation metrics specifically designed for assessing the quality of natural language explanations. We show that our model not only achieves precise EF predictions, but also generates explanations that are clinically relevant in a human-like language.}
\end{enumerate}

\section{Methods}
\label{sec:method}
\begin{figure}[t]
    \centering
    \includegraphics[width=\textwidth]{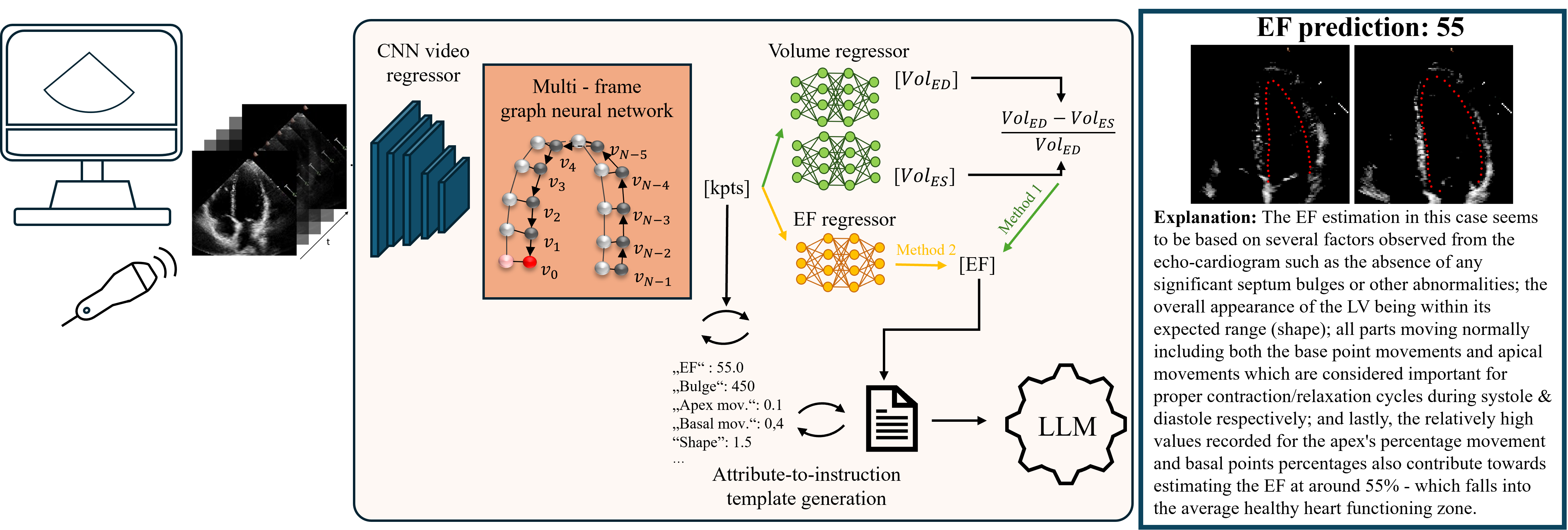}
    \caption{\textbf{Overview of the proposed pipeline}
    A US video is fed into a CNN video encoder that outputs a feature representation. The features are passed to a spatio-temporal GCN that returns keypoints for ED and ES. The keypoints serve as 1) input for MLPs that regress the LV volumes \textcolor{green}{(in green)} or the EF directly  \textcolor{orange}{(in orange)} and 2) computation of geometrical attributes that are converted into text snippets that can be parsed into an LLM. The LLM provides a human-like explanation for the EF.} 
    \label{fig:overview}
\end{figure}

Our approach introduces a streamlined pipeline that enriches a GCN-based EF prediction with clinically meaningful explanations. With echocardiography videos as input, a video encoder extracts feature representations, then fed to a spatio-temporal Graph Convolutional Network (GCN) which identifies anatomical keypoints. From these keypoints, the EF is predicted along with geometrical attributes essential for our text generator model, which produces natural language explanations of the EF predictions. An overview is shown in Fig.~\ref{fig:overview}.

\subsection{Multi-frame GCN Model for EF Prediction in a Single Pass}
\label{sec:gcn_model}

Central to our pipeline is the multi-frame Graph Convolutional Network (GCN) model, engineered to perform EF prediction and contour detection in a single integrated operation. Previous work \cite{ours2022miccai} designed a multi-task multi-frame model that derived the EF value directly from the input encoder. That approach resulted in EF and keypoints prediction being disentangled and less interpretable. Since manual EF computation relies solely on the contours, we modified the architecture to ensure that EF prediction follows the mathematical concept of the ratio between both keypoints volumes. By adding two volume regressors and then fusing the results, we base the model on prior knowledge about the dynamics rather than relying purely on black box predictions. We explored different levels of adding prior knowledge, either by directly regressing the EF from keypoints versus having two separate volume regression branches.

\subsection{Attribute generation}
\label{subsec:methods:attributeclassifier}

Beyond automatically predicting LV contours, our model also derives attributes that reflect structural changes and temporal deviations due to pathology or acquisition which affect the EF values. Based on clinical insights, we developed geometrical processing routines to compute attributes from LV contour points, capturing the intuition of cardiologists in EF assessment. This section outlines the attributes, grounded in clinical feedback, along with the computation.

\noindent\textbf{Septal bulge:} A septal bulge can be a morphological sign for early hypertensive heart disease \cite{Gaudron2016} and is an asymmetric, localized thickening of the basal-to-mid part of the inter-ventricular septum. It could be detected by calculating a wall thickness ratio over 1.4 \cite{marciniak2021septal}. We compute a bulge using the LV convexity which is the distance between a convex hull and the true contour. Ground truth contours were visually inspected while three manual thresholds were set to distinguish prominent bulges from undetected convexity calculated using OpenCV\footnote{www.opencv.org}.

\noindent\textbf{Segment motion:}
The 17-segment model\cite{american2002standardized} is widely used for regional wall motion analysis in multiple views. To simplify the process for 4CH-view, we divide the contour into 7 distinctive segments and calculate the segment movement direction relative to the overall motion as well as the vertical basal movement.

\noindent\textbf{Apex movement:} Foreshortening is a common problem in 2D echocardiography which results in underestimating the LV volume and inaccurate EF estimation. A foreshortened apex translates throughout the cycle, whereas a true apex almost remains at the same point. Following the approach of \cite{Smistad2020}, we compute the apex movement in the direction of the LV long axis. Based on the distribution in the dataset we set a threshold to indicate suspicious apex movement.

\noindent\textbf{Length-width ratio: }
A normal LV has a bullet-like shape. Cardiovascular diseases such as hypertension or heart failure may change the LV shape despite age and gender being also effecting factors. We decided to use the length-width ratio as a shape measure which is computed by dividing the apex-basal distance by the horizontal mid-septal distance. The length-width ratio is typically around \textbf{2}, as observed during our experiments, while in cases with dilated LV, resulting in a reduced length-width ratio.


\noindent\textbf{Sector intersection:} One important requirement for manual and automatic EF computation is to ensure full visibility of the LV within the ultrasound sector.  Therefore, we calculate the ratio of the intersection with the detected LV contours as a quality metric for the EF computation.

\noindent\textbf{Image quality:} 
Image quality will affect the visibility of the LV contours and the wall movement which influences the EF estimation.   
We calculate the intensity difference between the LV cavity and the myocardial wall as a measure of their contrast. A higher contrast indicates an improved visibility. However, this metric, though practical, does not cover all dimensions of image quality.


All of the aforementioned attributes are clustered based on their distribution in the annotated dataset. Thresholds were defined for each attribute to create text templates that could be fed into a language model.

\subsection{Generation of text explanations}
\label{subsec:methods:explanationgeneration}

The ability to translate complex geometrical and spatiotemporal attributes into understandable natural language explanations is critical to bridge the gap between advanced echocardiographic analysis and clinical practice. Our approach encompasses two primary phases: (1) converting the attributes into basic text sentences that describe the underlying clinical observations, and (2) refining these basic sentences into coherent natural language explanations suitable for clinical use, leveraging the capabilities of a Large Language Model (LLM).

{\bf From attribute values to LLM inputs. }
Computed attributes are numerical values that need to be converted into text tokens digestable for the LLM. The first phase involves translating the list of visual and geometric attributes into \textbf{basic sentences} by using thresholds. For instance, the numerical value \textit{bulge}$=500$ is translated to "A bulge value of 500 means that there is no bulge". 

{\bf Natural Language Refinement with LLM. }
In the second phase, we employ the LLaMA model~\cite{touvron2023llama}, a LLM variant, and train it for generating medical text specifically. We fine-tune LlaMA on a dataset with clinical explanations to ensure that the generated text aligns with clinical terminology and reasoning. To solve the limited availability of expert-generated explanations, we further implement some data augmentation strategies to enrich the training dataset.

{\bf Synthetic Explanations. } 
By adding prior clinical knowledge, we build more elaborated sentences from the basic sentences as synthetic expert explanations. These sentences articulate the clinical significance of each attribute in a structured format. For instance, an attribute indicating a significant septal bulge would be converted into a more elaborated sentence like "There is a large septal bulge, which may adversely affect the EF."

{\bf Data Augmentation through Self-instruction.}
We adopted a self-\\instruction method~\cite{wang2022self} using the GPT-4 model to augment a small initial dataset containing experts explanations. By feeding 5 expert explanations into GPT-4, along with a chain-of-thought prompt~\cite{Wei2022ChainOT} that includes examples of the input (basic sentences) and the desired output (expert explanation), we instruct GPT-4 to simulate medical expert explanations for novel sets of basic text. This use of chain-of-thought processing with GPT-4 effectively enlarges  our dataset towards a ten times expansion of the initial set.

This dual process ensures that our model not only accurately identifies the visual and geometrical attributes indicative of specific cardiac conditions, but also communicates findings in a way that clinicians can immediately interpret. 

\subsection{A novel metric to evaluate the EF explanation via LLMs}
\label{subsec:methods:metric_ef_text}
Evaluating unstructured text is crucial to identify errors in clinical LLM, yet human evaluation is time consuming and potentially subjective, highlighting the need for automated metrics. However, initial experiments indicated that even simple adversarial examples could deceive most of the existing metrics for sentence similarity. Given that the output of the proposed LLMs is unstructured text focused on key attributes, we designed a metric specifically aimed at assessing factual correctness. To accomplish this, we use the recently released Mistral model~\cite{jiang2023mistral}, another and faster LLM variant.
By creating nine targeted prompts with instructions and one-shot context for Mistral, we evaluate whether attributes appear in the text as positive (pathological), negative (normal), or unspecified. This allows a comparison between ground truth and prediction beyond mere textual similarity. In cases where an attribute is unspecified, its status is considered normal. We quantify the performance by reporting the accuracy, the count of true contradictions, hallucinations, and of missing attributes.

\section{Experiments}
\label{sec:experiments}
\subsection{Data}
\label{sec:data}
{\bf Dataset:} We use the EchoNet-Dynamic dataset~\cite{ouyang2020video}, which contains 10,030 echocardiography videos from healthy and pathological patients. Each video is annotated with 40 LV contour points, one basal and apex point at the end-diastolic (ED) and end-systolic (ES) frame, along with the EF. The training, validation and test splits provided by EchoNet are used for benchmarking. 

{\bf Annotations:} 
We trained the GCNs on the annotated keypoints from the EchoNet dataset, employing a multi-frame strategy in~\cite{ours2022miccai}. To simplify the processing, we selected ED and ES frames and sampled 14 evenly spaced intermediate frames from each video. For our experiments, ED and ES frames were assumed to be known as they can be computed from the ECG or automatically. A lack of Electronic Health Records (EHR) prompted us to enlist two cardiology experts who annotate a subset of the EchoNet data with video-text pairs. The experts watched the videos and provided text descriptions including an EF assessment and reasoning, focusing on attributes like LV shape, wall movement, and bulge presence. They were allowed to use different structures or description formats to ensure a diverse text reflection of real-world scenarios. 89 image-text pairs were generated for training, with additional 48 pairs designated for testing. 

\subsection{NLE evaluation metrics}
\label{sec:metrics}
We incorporated several different evaluation metrics to evaluate our LLM outputs from different aspects utilizing the advantages of each.
We exploit the \textbf{ClinicalBERT}~\cite{huang2019clinicalbert} and Sentence-based BERT models (\textbf{sBERT})~\cite{rasmy2021med} that generate single embeddings either per word or per sentence followed by cosine similarity and are pre-trained on clinical texts. Additional text similarity models are provided in the suppl. material.
The Mistral score was introduced to evaluate explanations against specific clinical attributes, leveraging a recently developed \textbf{Mistral LLM}~\cite{jiang2023mistral} tailored for this purpose. This custom metric (sec. \ref{subsec:methods:metric_ef_text}), aims to provide a more nuanced assessment of the clinical relevance and accuracy of the explanations generated. In addition to accuracy metrics, we also used the Flesch Reading Ease score to measure contextual richness of the explanation (a lower Flesch score means contextually richer text).

\subsection{EF predictions with generated explanations}
\label{sec:predictions}

{\bf Implementation:} Our GCN model uses a ResNet-3D-18 as video encoder backend, optimizing for the efficient processing of echocardiography videos. GCN model training focuses on accurately predicting LV keypoints, which are crucial for the subsequent estimation of EF and the generation of explanations. For the NLE component, we train the LLaMA model to generate clinically relevant explanations based on attributes derived from the GCN output. 
LLM training included a low-rank adaptation on the LLaMA-1 model through 8-bit quantization, with a learning rate of $0.0003$ and a batch size of 32. Training was performed for 5 hours on two A6000 GPUs, each equipped with 48GB memory. 
The end-to-end inference pipeline ensures a seamless transition from raw video data to EF predictions accompanied by understandable explanations. For this experiment, the GCN with the lowest mean absolute error (MAE) was used. Detailed implementation specifics are available on GitHub for reproducibility\footnote{https://github.com/guybenyosef/EchoNarrator}.

\begin{figure}[t]
    \centering
    \includegraphics[width=0.97\textwidth]{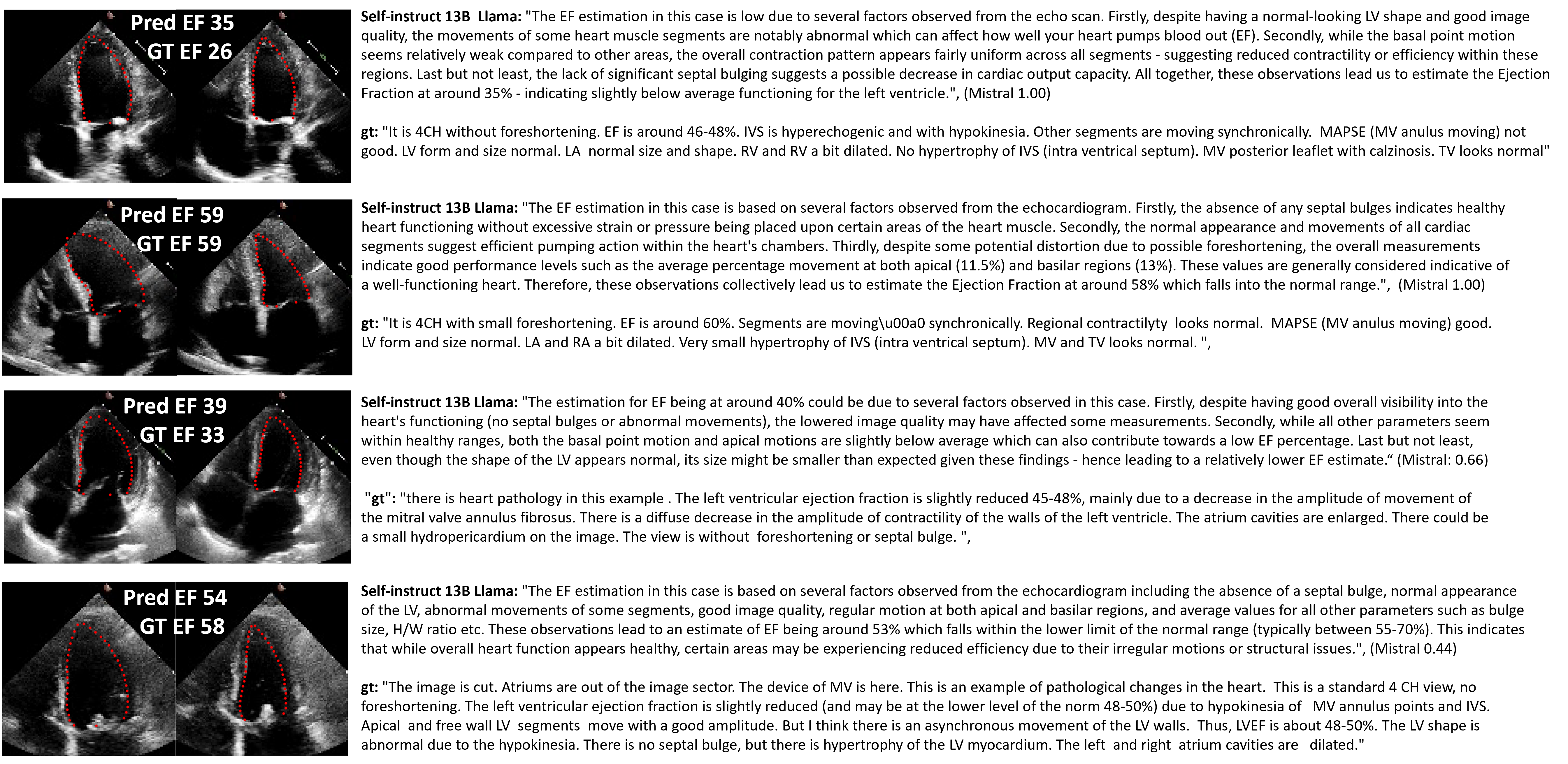}
    \caption{(Zoom in for optimal view) LV contour estimation, EF prediction and its text explanation as provided by the NLE-EF-13B self-instruct on EchoNet test examples.} 
    \label{fig:plots}
\end{figure}

{\bf Results:} In evaluating our end-to-end inference system, we focus on both the EF prediction accuracy and the quality of the generated explanations. While competitors exist for EF prediction, our approach is unique in integrating NLE, setting a benchmark in the field. 
Tab.~\ref{tab:ef} lists the details of our comparison with previous models using the Dice score, the mean keypoint error (MKE) and the accuracy of the prediction of EF. We show that our single task NLE GCN can reach a lower MKE than the EchoGraph while maintaining the same EF accuracy. Predicting from volumes (Vol) versus predicting directly from keypoints (EF) performed similarly well.  
For NLE prediction, we evaluate the coherence and clinical relevance of the generated explanations quantitatively (Tab.~\ref{tab:nle}) and qualitatively (Fig.~\ref{fig:plots}). In Tab.~\ref{tab:nle}, we added a prediction of the LLaVA-Med model~\cite{li2024llava} when we input an image showing ED and ES frames, followed by the instruction to explain the EF (see suppl. material for details). Although different from our model in the design and goal, it seems to be the closest in providing baseline text explanations.

{\bf Analysis:} 
Our results demonstrate notable accuracy in EF prediction (Tab.~\ref{tab:ef}) combined with the generation of clinically meaningful explanations. Compared to competitors in EF predictions, the scores in Tab.~\ref{tab:nle} further show that our system not only achieves good EF estimation, but also introduces the capability of generating accurate, complete, and human-like explanations. Tab.~\ref{tab:nle} also shows that predicted explanations are better from simple baselines. The enhancement of NLE prediction in our model can be attributed to the use of synthetic data and data augmentation techniques (Sec.~\ref{subsec:methods:explanationgeneration}). The utilization of the self-instruction by using Chain-of-Thought in GPT4 further refines the model's capability to generate plausible and contextually more rich explanations that meet clinical expectations, as shown by its lower scores in the Reading Ease metric.

\begin{table}
\centering
\resizebox{\textwidth}{!}{%
\begin{tabular}{|l|c|c|c|c|c|l|}
\hline
Model & Frames & Dice (\%) & MKE (px) & EF MAE & Cycle & Explainable \\ \hline
EchoNet~\cite{ouyang2020video} & 32 & $91.7 \pm 4.2$ & $2.5\pm1.2$ & 4.22 & Single & No \\ 
EchoGraphs~\cite{ours2022miccai} & 16 & $90.3\pm4.3$ & $2.7\pm1.5$ & 4.01 & Single & No \\ 
EchoCoTr-S~\cite{rand2022miccai} & 36 & N/A & N/A &3.95  & Multi & No \\
GEMTrans~\cite{Mokhtari2023} & 16  & N/A & N/A &4.15  & Multi & Heatmaps \\
NLE EF GCN (Ours) & 16 & $91.5\pm4.3$  & $2.4\pm1.1$ & 4.00 & Single & Text \\
NLE Vol GCN (Ours) & 16 & $91.4\pm4.4$ & $2.4\pm 1.1$ & 4.05 & Single & Text \\ 
\hline
\end{tabular}%
}
\caption{Segmentation and EF accuracy for different methods evaluated on EchoNet testset (n=1264) and annotated ED and ES frames (MKE = mean L1 keypoint pixel error). \textit{Multi} refers to whole videos whereas \textit{single} refers to one cycle or ED to ES.
}
\label{tab:ef}
\end{table}

\begin{table}[t]
\begin{tabular}{|l|c|c|c|c|c|c|c|}
\hline
Model          & mistral acc$\uparrow$& halluc.$\downarrow$ &contradict.$\downarrow$& missing$\downarrow$& cbert$\uparrow$ & msbert$\uparrow$  & Flesch$\downarrow$ \\
\hline
random\footnotemark                &   0.45      &                    &                &             &                    &                &  \\
LLaVA-Med        &  0.67     &        1.49  &     0.87       &               1.49        &           0.92         &     0.95              &  54.92 \\
NLE-EF-13B      & 0.77                         & 0.54       & 1.02                  & 1.50           & 0.93                     & 0.96                          & 58.59                        \\
NLE-EF-7B       & 0.77                          & 0.65          & 0.67               & 1.44                & 0.94                     & 0.95                    & 57.20                        \\
NLE-EF-13B self-i & 0.77                    & 0.44      & 1.13                    & 1.65                     & 0.91                     & 0.95                    & 30.91                        \\
NLE-EF-7B self-i & 0.80                         & 0.50       & 0.90                & 1.33           & 0.92                     & 0.95                            & 27.42                        \\
\hline
\end{tabular}
\caption{Evaluation of the NLE-EF versions on 48 samples from the EchoNet test set using different averaged metrics such as mistral score accuracy, average number of mistral contradictions, hallucinations and missing attributes, cbert score (clinical BERT), msbert score (sentence medsBERT), and Flesch reading ease.
}
\label{tab:nle}
\end{table}
\footnotetext{{Means random answers to the attributes from sec.\ref{subsec:methods:attributeclassifier}} }

\section{Discussion and conclusion}
\label{sec:discussion}
Our method introduces an innovative approach by leveraging a GCN and a LLM (LLaMA) to provide LV contours and EF along with geometrical features and a text explanation.  
The main contribution is the integration of cardiac features (potentially less intuitive for humans) derived from a vision model with an LLM that translates these features into explanatory text. Considering that the primary focus was on the effective combination of these components to enhance interpretability, architectural choices were based on experiments. Leveraging synthetic and augmented data can improve interpretability without compromising prediction accuracy. This balance is vital for wider clinical adoption, where the clarity of the explanations is as important as the accuracy. Our evaluation with another LLM aims to increase sensitivity to contradictions while configurations and prompt design need to be considered carefully. Despite notable successes, we acknowledge limitations such as a relatively small dataset, noisy labels and prompts, which could affect our findings' robustness and generalizability. We incorporated six widely used LV attributes, but clinical feedback suggested extending this to include the right side of the heart. Although GCN and LLM pre-training add more data implicitly, an extension of the dataset, including more attributes, and a clinical evaluation will be future work. The proposed method facilitates AI-assisted diagnosis, reporting, and education by providing cardiologists an accurate visual output with human-readable explanations.

\begin{credits}
\subsubsection{\ackname} 
This work was financially supported by the Research Council of Norway (RCN), through an innovation project (EchoAI, 313756) and its Centre for Research-based Innovation (Visual Intelligence, 309439). We thank E. Steen, J. Sprem and S. A. Aase for their guidance and feedback.
\subsubsection{\discintname}
The authors declare no further competing interests. A patent has been filed for the methods and technologies described in this document.
\end{credits}
%
%
\bibliographystyle{splncs04}
\bibliography{main}

\appendix
\section*{Loss Functions for EF, Keypoints, and NLE Predictions}

Our model optimizes a combination of different loss functions to ensure accurate EF predictions and high-quality clinically relevant explanations.

\subsection*{1. Regression Loss for EF Prediction}

For the prediction of EF, we use a standard mean squared error (MSE) loss to minimize the difference between the predicted EF ($\hat{EF}$) and the true EF ($EF$). This can be defined as follows.

\[
\mathcal{L}_{\text{EF}} = \frac{1}{N} \sum_{i=1}^{N} (\hat{EF}_i - EF_i)^2
\]

where $N$ is the number of samples in the batch.

\subsection*{2. Geometrical Keypoint Loss}

To ensure accurate prediction of the LV contour, we utilize a loss function based on the Euclidean distance between the predicted keypoints ($\hat{k}_i$) and the true keypoints ($k_i$) for each frame:

\[
\mathcal{L}_{\text{keypoints}} = \frac{1}{N} \sum_{i=1}^{N} \sum_{j=1}^{M} \|\hat{k}_{ij} - k_{ij}\|^2
\]

where $M$ is the number of keypoints in each frame.

\subsection*{3. Text Explanation Loss}

For generating coherent and accurate text explanations, we train the LLM using a cross-entropy loss, which measures the divergence between the predicted token distribution and the ground-truth token distribution. This is given by:

\[
\mathcal{L}_{\text{text}} = - \frac{1}{T} \sum_{t=1}^{T} \log P(y_t | y_{1:t-1}, \mathbf{h}_{t-1})
\]

where $T$ is the total number of tokens in the explanation, $y_t$ is the ground truth token at the time step $t$, and $P(y_t | y_{1:t-1}, \mathbf{h}_{t-1})$ is the predicted probability of the token based on the previous tokens $y_{1:t-1}$ and the hidden state $\mathbf{h}_{t-1}$ of the top transformer layer.

The hidden state $\mathbf{h}_{t-1}$ represents the contextual information (or embedding) derived from all tokens up to the time step $t-1$ and is used to generate the probability distribution to predict $y_t$.

\subsection*{4. Total loss}

The overall loss function is a weighted combination of the individual losses described above.

\[
\mathcal{L}_{\text{total}} = \lambda_1 \mathcal{L}_{\text{EF}} + \lambda_2 \mathcal{L}_{\text{keypoints}} + \lambda_3 \mathcal{L}_{\text{text}}
\]

where $\lambda_1$, $\lambda_2$, and $\lambda_3$ are hyperparameters that control the relative importance of each loss term.

\end{document}


\title{Supplementary material}
\author{}
\institute{}
\maketitle
\vspace{-10mm}
\section{Implementation details}

\begin{table}[]
\centering

\begin{tabular}{@{}p{2.1cm}lllcc@{}}
Model &Exp. & Backbone &  GCN layers & EF layers  & Batch size \\ \hline 
 GCN sector &sector segm. & MobileNet2 & [4,8,8,16,16,32,32,48] &  - & 128   \\ 
NLE EF GCN &vol from kpts& r3d\_18    & [16,32,32,48] &    [16,32,32,48]  & 12 \\ 
NLE Vol GCN&ef from kpts& r3d\_18   & [16,32,32,48] &  [16,32,32,48] & 12  \\ 
\end{tabular}
\label{tab:networks}
\vspace{1mm}
\caption{Overview of different experiment configurations. Two multi GCN configurations are listed and also a simple sector GCN model that is trained on a subset of the training data to segment the outline of the triangular EchoNet sector in the US image.}
\end{table}
\vspace{-15mm}

\begin{table}[]
\centering
\begin{tabular}{@{}p{3.5cm}p{9cm}@{}}
Parameter & Value \\ \hline 
 Public dataset & EchoNet \\
 Number of images & 10.030 videos excluding 90 irregular keypoint annotations \\
 Input image size & [3,112,112,16] \\ 
 Optimization & Optimizer Adam, learning rate $10^{-4}$\\ 
 Pre-training & Kinect-400 (Multi-frame) \\ 
 Augmentations & aligned with [23] for multiple frames
\end{tabular}
\label{tab:networks}
\caption{Details on data pre-training and hyperparameters}
\end{table}
\vspace{-5mm}
\section{Additional metrics}
\vspace{-5mm}
\begin{table}
\begin{tabular}{ccccccccccc}
Model & bert & sbert & biobert & simcse & medsbert & rouge & rougeL & bleu & wcount \\
\hline
LlaVA-Med   & 0.81 & 0.49  & 0.53           & 0.69   & 0.95     & 0.25  & 0.15   & 0.01 & 541        \\
NLE-EF-13B           & 0.88 & 0.56  & 0.63           & 0.75   & 0.96     & 0.26   & 0.15   & 0.00 & 194         \\
NLE-EF-7B             & 0.88 & 0.56  & 0.63            & 0.75   & 0.95     & 0.25   & 0.15   & 0.00 & 207         \\
NLE-EF-13B self-instr    & 0.84 & 0.54  & 0.59            & 0.75   & 0.95     & 0.21   & 0.12   & 0.00 & 541         \\
NLE-EF-7B self-instr    & 0.85 & 0.57  & 0.62           & 0.75   & 0.95     & 0.19   & 0.10   & 0.00 & 622         \\
\end{tabular}
\caption{Additional metrics for the different NLE approaches. BERT, BioBERT, SimCSE are based on words, whereas sBERT,medsbert are on sentence level. BioBERT and medsbert are trained on clinical data. SimCSE is targeted for detecting contradictions. Rouge and BLEU are conventional NLP scores that are also dependent on the text length. Word count (wcount) accounts for the diversity of the different approaches.}
\end{table}

\vspace{-15mm}
\section{Prompt examples}
\tiny
\begin{lstlisting}[frame=single] 
Example 1: Llama input prompt
{"instruction": 
"Explain why the ejection fraction is estimated as 69%. ",
"input": "In the echocardiography image, it is measured that there is no septal bulge 
present, the shape of the left ventricle looks normal, the movement of the segments 
is normal, the image quality is reduced, the basal points move normal, the left 
ventricle is cut, and the apex moves with a large amplitude. There might be 
foreshortening. The bulge value is 417, the height over width value is 2.00, the
segment movement is normal, the apex moves 13.61%, the basal points move  by 7.28%,
contrast is 4.76.",
"response": "The EF estimation in this case seems to be based 
on several factors observed from ....}
\end{lstlisting} 
\normalsize

\tiny
\begin{lstlisting}[frame=single]
Example 2: Prompt for the attribute Bulge for the Mistral model
<s> [INST] I have the following text that describes an image and I want you to answer
some questions about it by selecting one from different options. [/INST] Sure let me
help you with that, what is the text and the question </s> <s>[INST] This is the 
text: There is a large bulge. Does the text mention any kind of bulge? Pick ONE 
final answer out of: [1/bulge], [2/not specified in the text] or [3/none]. [/INST]
Final answer: [1/bulge] \n\n Explanation: The text states that there is a large bulge
present in the image, therefor it is present. </s><s> [INST]  Great. Now do the 
same task for the following text: TEXT_TO_INSERT Does the text mention any kind 
of a bulge? Pick ONE final answer out of: [1/bulge], [2/not specified in the text]
or [3/none]. And please answer in the format:
Final answer: [option] \n\n Explanation: text [/INST]
\end{lstlisting} 
\normalsize
\begin{figure}
    \includegraphics[width=1.04\textwidth]{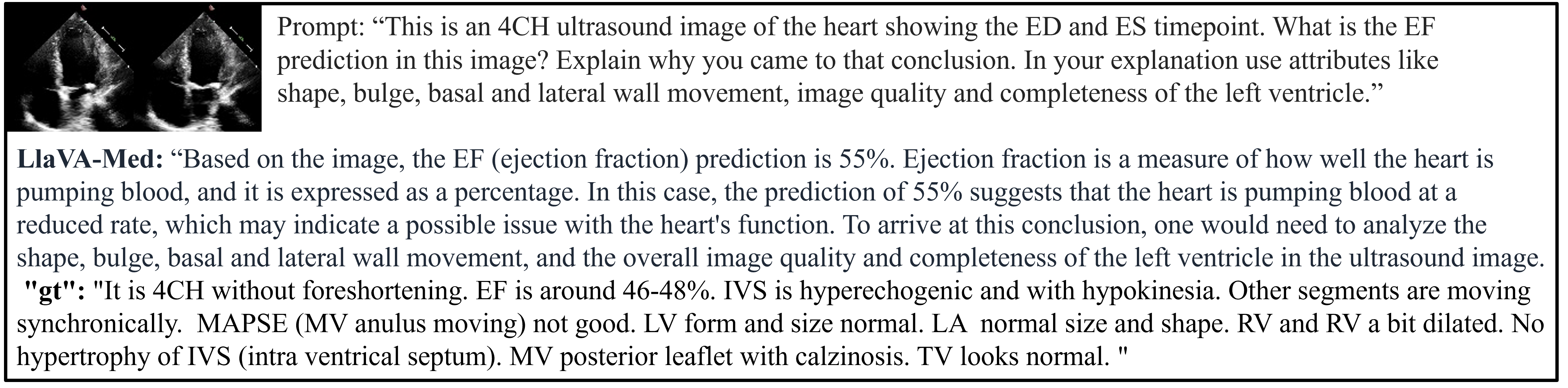}
    \caption{Prompt example used in LlaVA-Med. To compare with the NLE-EF output, we present here the same most-left example shown in Fig.2 in the main paper. } 
    \label{fig:llava}
\end{figure}
\vspace{-5mm}

\section{Attributes}
\vspace{-10mm}
\begin{figure}[]
    \centering
    \includegraphics[width=1.2\textwidth]{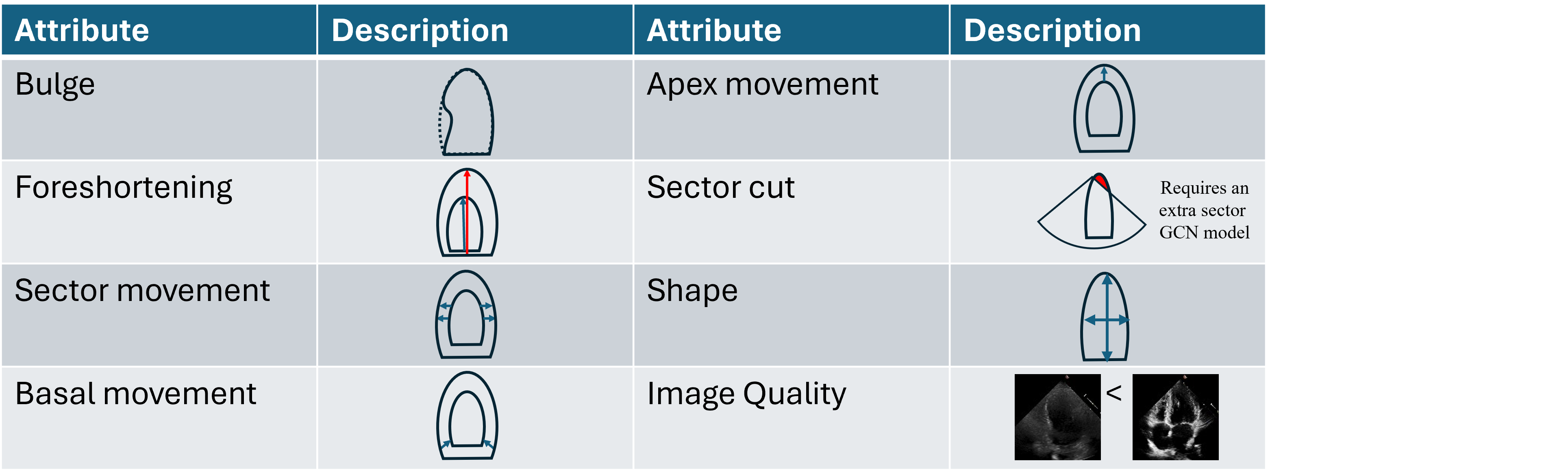}
    \caption{Overview on attributes: Illustrative description of the different attributes used as input for the text model. Image Quality and sector cut are based on the image.} 
    \label{fig:single}
\end{figure}

